\documentclass[preprint,12pt]{elsarticle}

\usepackage{amssymb}

\usepackage{booktabs}
\usepackage{multirow}
\usepackage{xspace}
\usepackage{xcolor}
\newcommand{\tuple}[1]{\ensuremath{\langle #1\rangle}}
\newcommand{\Omri}[1]{}
\newcommand{\Carmel}[1]{}
\newcommand{\Roni}[1]{}
\newcommand{\shortcite}[1]{\cite{#1}}
\newcommand{\gtv}[1]{\ensuremath{\textit{G2V}}\xspace}
\newcommand{\fgtv}[1]{\ensuremath{\textit{FG2V}}\xspace}
\newcommand{\kaduri}[1]{\ensuremath{\textit{KBS}}\xspace}
\newcommand{\mapfgas}[1]{\ensuremath{\textit{MAG}}\xspace}

\usepackage{colortbl}
\definecolor{Gray}{gray}{0.9}

\journal{Arxiv}

\begin{document}

\begin{frontmatter}



\title{Algorithm Selection for Optimal Multi-Agent Path Finding via Graph Embedding}


\author[bgu]{Carmel Shabalin}
\author[bgu]{Omri Kaduri}
\author[bgu]{Roni Stern}
\affiliation[bgu]{organization={Ben Gurion University of the Negev},
            addressline={Beer Sheva blvd 1},
            city={Beer Sheva},
            postcode={84105},
            country={Israel}}

\begin{abstract}
Multi-agent path finding (MAPF) is the problem of finding paths for multiple agents such that they do not collide. 
This problem manifests in numerous real-world applications such as controlling  transportation robots in automated warehouses, 
moving characters in video games, and coordinating self-driving cars in intersections. 
Finding optimal solutions to MAPF is NP-Hard, yet modern optimal solvers can scale to hundreds of agents and even thousands in some cases. 
Different solvers employ different approaches, and there is no single state-of-the-art approach for all problems. 
Furthermore, there are no clear, provable, guidelines for choosing when each optimal MAPF solver to use. 
Prior work employed Algorithm Selection (AS) techniques to learn such guidelines from past data. 
A major challenge when employing AS for choosing an optimal MAPF algorithm is how to encode the given MAPF problem. 
Prior work either used hand-crafted features or an image representation of the problem. 
We explore graph-based encodings of the MAPF problem and show how they can be used on-the-fly with a modern graph embedding algorithm called FEATHER. 
Then, we show how this encoding can be effectively joined with existing encodings, resulting in a novel AS method we call MAPF Algorithm selection via Graph embedding (\mapfgas\ ). An extensive experimental evaluation of \mapfgas\ on several MAPF algorithm selection tasks reveals that it is either on-par or significantly better than existing methods. 
\end{abstract}




\end{frontmatter}


\section{Introduction}
\label{scn:Intro}

Multi-Agent Path Finding (MAPF) is the problem of finding paths for a group of agents, moving each agent from its initial location to a designated target location. 
The main constraint in MAPF is that the agents must not collide. 
MAPF has received significant interest recently in the scientific community and in industry. 
Practical applications include controlling transportation robots in automated warehouses~\cite{wurman2008coordinating}, 
coordinating self-driving vehicles in intersections~\cite{li2023intersection}, 
pipe-routing in gas plans~\cite{belov2020from}, 
embedding virtual network requests in computer networks~\cite{zheng2023priority}, and moving characters in video games~\cite{ma2017feasibility}. 
Finding an optimal solution to MAPF with respect to various optimization criteria is known to be NP Hard~\cite{anOptimization2010surynek,structure2013yu}. 
Nevertheless, a range of practical algorithms that guarantee optimality exists~\cite{kornhauser1984pebble,surynek2009novel}. 
It has been shown that these algorithms are able to find optimal solutions to MAPF problems with more than 100 agents with less than one minute of runtime~\cite{li2021pairwise}. 

Different optimal MAPF algorithms employ different problem-solving techniques. For example, EPEA*~\cite{goldenberg2014enhanced} employs a heuristic search technique, BCP~\cite{lam2022branch} uses optimization techniques, and SAT-MDD~\cite{surynek2016efficient} compiles MAPF to Boolean Satisfiability. 
Correspondingly, different solvers work best for different MAPF instances, and no algorithm has emerged to dominate all others. 
The variance in performance can be great, where some algorithms perform poorly in some instances but significantly outperform all others in other instances. 
On a recently performed extensive comparison of 5 optimal MAPF algorithms~\cite{kaduri2021experimental}, it was shown that even the least effective algorithm had some grids in which it was able to solve instances with 4 times more agents than all other evaluated algorithms. 
Thus, developing methods for selecting the best optimal MAPF search algorithm for a given instance is a worthwhile endeavor. 


The problem of determining which algorithm from a given set of algorithms is expected to perform best on a given problem instance is known as the 
Algorithm Selection (AS) problem~\cite{rice1976algorithm,kerschke2019automated}.\footnote{Technically, this problem is called the \emph{per-instance AS} problem.}
Several recent works have developed AS techniques for optimal MAPF~\cite{kaduri2020algorithm,ren2021mapfast,alkazzi2022mapfaster}. 
They used supervised learning to
train a classifier that chooses the best optimal MAPF algorithm for a given MAPF instance, from a portfolio of optimal MAPF algorithms that include EPEA*~\cite{goldenberg2014enhanced}, ICTS~\cite{sharon2013increasing}, SAT-MDD~\cite{surynek2016efficient}, CBSH~\cite{felner2018adding}, and Lazy CBS~\cite{gange2019lazy}.  
A key challenge faced by all prior work on AS for optimal MAPF is how to encode a given MAPF instance, as an input to the supervised learner. 
Kaduri et al.~\cite{kaduri2020algorithm} proposed to encode a MAPF problem as a vector of hand-crafted MAPF-specific features~\cite{kaduri2020algorithm}, such as the average length of the shortest path between the start and goal locations. 
Others~\cite{alkazzi2022mapfaster,ren2021mapfast} proposed to cast a MAPF instance as an image.

Ren et al.~\cite{ren2021mapfast} mentioned the possibility of encoding a MAPF instance as a graph comprising the shortest paths of the different agents. However, they consider this method to be ``not a deployable algorithm selector in any reasonable sense,'' since it could not be used on any MAPF problem not observed during training. 
\textbf{The first contribution} is a novel encoding of the MAPF instance called \fgtv\ , that fully utilizes the power of graph embedding algorithms by encoding the entire graph, augmented with artificial edges marking the start and target vertices of every agent. 
This embedding is done using FEATHER~\cite{rozemberczki2020characteristic}, a modern graph embedding algorithm that can be used on-the-fly on any given graph. 
The resulting embedding yields superior results in most cases, but not always.

In general, neither \fgtv{} nor any of the existing encodings completely captures all aspects of the encoded MAPF instance. 
\textbf{The second contribution} is a simple method for integrating multiple encodings, which enables a more comprehensive encoding of the problem. This method can be used with different embeddings in a seamless manner. The resulting AS method is called \mapfgas\ . 
\textbf{Our third contribution} is a comprehensive evaluation of \mapfgas\ on a standard benchmark over three different AS tasks, which differ in how similar are the train and test instances. Our results show that \mapfgas\ is superior to baseline methods, demonstrating \textbf{the first effective use of graph embedding for solving MAPF problems}. All our code and datasets will be made publicly available to the community. 




\section{Background}
\label{scn:Background}

\noindent For completeness, we provide here a brief background on MAPF, AS for MAPF, and graph embedding. 

\subsection{MAPF}
A MAPF problem instance is defined by a tuple $\tuple{k,G,s,t}$ 
where 
$k$ is the number of agents, 
$G = (V;E)$ is an undirected graph, 
$s: [1\ldots, k]\rightarrow V$ maps an agent to its source vertex, 
and $t: [1,\ldots, k]\rightarrow V$ maps an agent to its target vertex. 
Initially, each agent is in its source vertex.
In every time step, each agent either waits
in its current vertex or moves to one of the vertices adjacent to it. 
A single-agent plan for agent $i$ is a sequence of move/wait actions that moves $i$ from $s(i)$ to $t(i)$. 
A solution to a MAPF instance is a set of single-agent plans, one for each agent, such that the agents do not collide with each other. 
In the classical MAPF problem, the cost of a solution is either the sum of actions in all single-agent plans
or the number of actions in the longest single-agent plan. The former cost function is known as sum-of-costs (SOC) and the latter is known as makespan. 
We measured solution cost in terms of SOC in our experiments, but most of our contributions are agnostic to the chosen cost function.   
MAPF extensions that consider non-unit action costs, large agents, and continuous time have been explored~\cite{atzmon2020generalizing,andreychuk2022multi,li2019multi}. In this work, we focus on classical MAPF.

Optimal classical MAPF algorithms are classical MAPF algorithms that are guaranteed to return optimal solutions according to a predefined solution cost function. 
Different optimal algorithms have been proposed for solving classical MAPF problems. 
Prime examples are EPEA*~\cite{goldenberg2014enhanced}, SAT-MDD~\cite{anOptimization2010surynek}, CBS~\cite{sharon2015conflict,felner2018adding} and its many extensions, BCP~\cite{lam2022branch}, Lazy CBS~\cite{gange2019lazy}, and ICTS~\cite{sharon2013increasing}. These algorithms apply a range of techniques: some use heuristic search on dedicated state spaces, other compile the problem to Boolean Satisfiability (SAT) and call an off-the-shelf SAT solver, while others borrow ideas from constraint programming. 
No algorithm fully dominates the other, and different  problems are solved best with different algorithms. This raised the need for an automated way to select which optimal MAPF solver to use for a given problem.

\subsection{AS for MAPF}

Previously proposed AS methods for MAPF followed a standard supervised learning paradigm, 
and differ mainly in the type of features they extract. 
Sigurdson et al.~\cite{sigurdson2019automatic} and Ren et al.~\cite{ren2021mapfast} mapped a given MAPF problem to an image and extracted image-based features with a Convolutional Neural Networks (CNN). 
Kaduri, Boyarski, and Stern~\cite{kaduri2020algorithm} proposed a set of hand-crafted, MAPF-specific features, such as the number of agents divided
by the number of unblocked cells in the grid. 
We refer to these features as the \kaduri\ features. 
As noted above, Ren et al.~\cite{ren2021mapfast} also explored the potential of using features that are based on mapping the given MAPF problem to a graph and using a \emph{graph embedding} method. 
While their exploration showed that graph embedding can provide useful features for MAPF AS, they do not provide a practical method to use it since it could not be used on any MAPF problem not observed during training. 
We overcome this limitation in our work.  


Kaduri et al.~\cite{kaduri2021experimental} distinguished between three types of AS problems for MAPF: 
(1) in-grid AS, (2) in-grid-type, and (3) between-grid-type. 
In-grid AS means the train and test problems are all from the same underlying graph. 
In-grid-type means the train and test problems are on different grids, but their grids have similar topologies. 
Between-grid-type means that graphs in the train and test problems are completely different, e.g., training on maze-like graphs and testing on graphs that represent roadmaps in a city. Most prior work has focused on the in-grid AS problem, which is, of course, significantly easier.







\subsection{Graph Embedding Algorithms}

Node embedding methods  are algorithms for encoding a node in a graph into a low-dimensional continuous vector~\cite{goyal2018graph}. 
Similarly, graph embedding methods are algorithms for encoding an entire graph $G$ into a low-dimensional continuous vector~\cite{goyal2018graph}, referred to as the \emph{embedding} of $G$. 
Ideally, graphs with similar structure will have embedding that are close in terms of their Euclidean distance. 
Graph embeddings have proven to be useful features for various machine learning tasks such as classification~\cite{you2020cross}.

Graph2Vec~\cite{narayanan2017graph2vec} is a neural graph embedding algorithm that accepts a set of graphs and outputs an embedding for each graph by analyzing local neighborhood of their nodes. It runs stochastic gradient descent to optimize a loss function that ensures similar graphs in the input set of graphs will have a similar embedding. Notably, it cannot be effectively used on other graphs without re-training, and thus its usefulness for our purposes is limited. 
FEATHER~\cite{rozemberczki2020characteristic} is a recently proposed algorithm that serves as both a node embedding and a graph embedding algorithm. 
Its embedding is based on the likelihood of reaching each node in a random walk. 
Unlike Graph2Vec, it does not require an optimization step, and can reasonably used to embed a single graph.  
The embeddings created by FEATHER have shown to be effective in node-level and graph-level machine learning tasks, such as classifying fake users in a social network and link prediction.
FEATHER has an additional positive property that it describes isomorphic graphs with the same representation and exhibits robustness to data corruption. 
\section{Method}
\label{scn:Methods}

In this section, we describe our AS method for choosing optimal classical MAPF algorithms called MAPF Algorithm selection via Graph embedding (\mapfgas\ ). 
\mapfgas\ works as follows. 
First, it encodes the given MAPF problem as a graph. 
Then, it creates an embedding of the resulting graph with FEATHER. 
The resulting vector is added to a set of previously proposed MAPF-specific features extracted from the given MAPF problem, 
and used to solve our AS problem using supervised machine learning. 
Next, we describe each of these steps in more detail.

\subsection{Encoding MAPF as a Graph}

We consider two ways to encode a MAPF problem $\Pi=\tuple{k,G,s,t}$ as a graph. 
The first encoding method, called \gtv\ , was previously proposed by Ren et al.~\cite{ren2021mapfast}.
\gtv\ encodes $\Pi$ by extracting from $G$ that induced subgraph that includes only nodes that are on a shortest path between the source and the target of an agent.
In more detail, G2V computes for each agent the shortest path from its source to its target while ignoring all other agents. 
Then, we construct a graph that contains only the nodes on these shortest paths, adding an edge between any pair of nodes on these shortest paths that have an edge between them in the original MAPF problem. 
Note that the resulting graph may be disconnected. 

\begin{figure}
    \centering
    \includegraphics[width=\columnwidth]{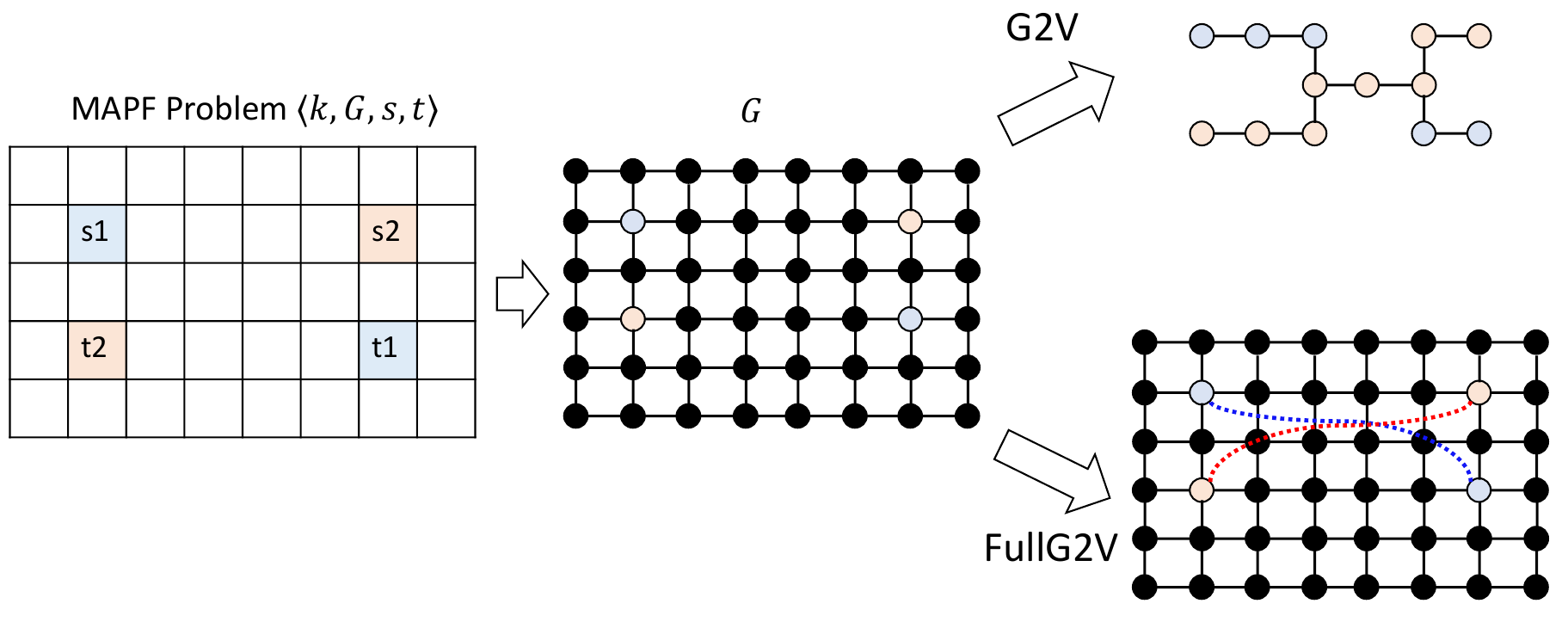}
    \caption{An example of the \gtv\ and \fgtv\ encoding methods.} 
    \label{fig:graph-encodings}
\end{figure}

The benefit of G2V is that the resulting graphs are significantly smaller than $G$. 
However, these graphs lose information: they do not consider regions on $G$ that are not on the agents shortest paths. 
These regions may be important to consider when the corresponding MAPF problem is particularly difficult and agents must move away from their shortest paths. 
To address this limitation of \gtv, we propose \emph{FullG2V} (\fgtv\ ), which uses the entire graph $G$ to encode the given MAPF problem. 
To include details about the MAPF problem beyond $G$, the graph outputted by \fgtv\ includes an artificial edge for every agent between its source and target. 
Formally, for a MAPF problem $\tuple{k,G=(V,E),s,t}$ \fgtv\ outputs the graph $G'=(V',E')$ where 
$V'=V$ and $E'=E\cup \{(s(i),t(i))\}_{i=1}^k$. Figure~\ref{fig:graph-encodings} illustrates \gtv\ and \fgtv\ on a simple MAPF problem on a 4-neighborhood grid. 

\subsection{Embedding the Graph}

To embed the encoded graph into a vector space, we used the FEATHER algorithm~\cite{rozemberczki2020characteristic}. 
FEATHER creates a graph embedding by first embedding each of the graph nodes and then pooling the resulting vectors to a single vector of size 500. 
Unlike other graph embedding techniques, it does not require a-priori training. 
Consequently, the features extracted using FEATHER can be extracted and used in a meaningful way even for graphs created for MAPF problems that are not in the training set. 
This is key to allowing graph embedding features to be used for optimal MAPF AS methods. 

Note that the default pooling method in FEATHER is ``mean''. 
This means the graph embedding is created by taking the mean over its constituent node embeddings.
We observed that using mean pooling did not perform well in our context, i.e., led to poor classification results when used as features. 
The reason for this is that mean pooling diminishes the impact of the artificial edges added between the source and target of each agent. Thus, MAPF problems on the same graph yielded very similar embeddings. 
To overcome this, we configured FEATHER to use ``max'' pooling, which emphasizes small differences between graphs created from MAPF problems on the same grid. This yielded significantly better results when  training the AS classification model.

\subsection{Feature Extraction and Learning}

\begin{figure}
    \centering
    \includegraphics[width=\columnwidth]{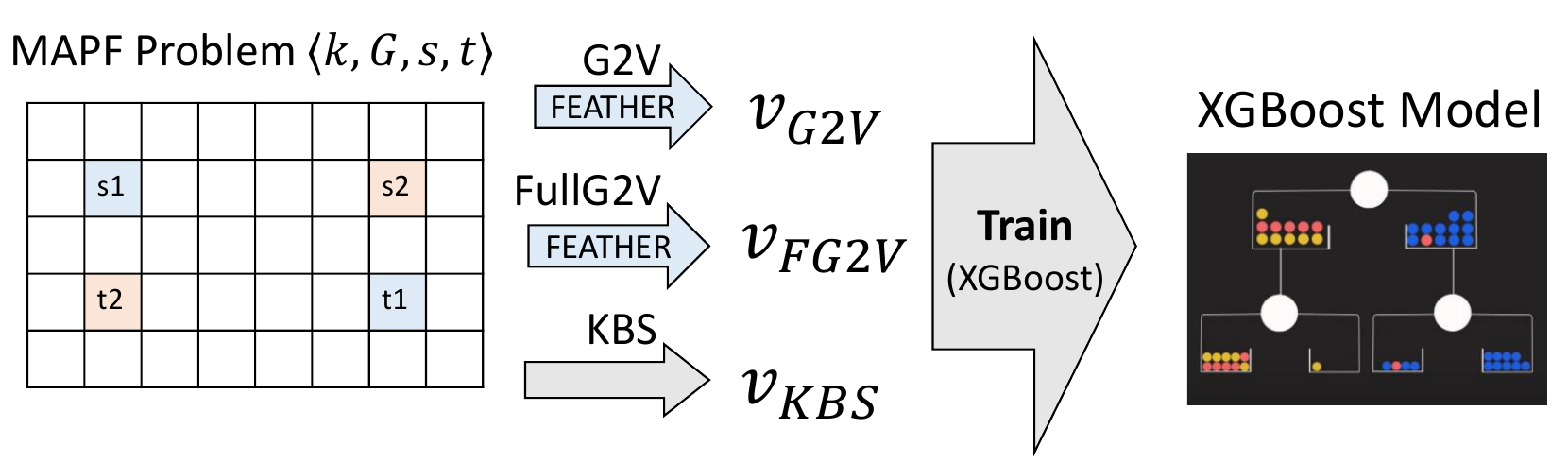}
    \caption{Diagram of the \mapfgas\ feature extraction and training process.}
    \label{fig:mgs-features}
\end{figure}

For a given MAPF problem $\Pi$, \mapfgas\ uses \gtv\ and \fgtv\ to create two graphs $G_\gtv\ $ and $G_\fgtv\ $ that encoding $\Pi$. 
Then, it creates two graph embeddings $v_\gtv\ $ and $v_\fgtv\ $ by applying FEATHER on these graphs. 
It also creates a vector $v_\kaduri\ $ by extracting all the MAPF-specific features proposed by Kaduri et al.~\cite{kaduri2020algorithm}. 
Finally, \mapfgas\ concatenates 
$v_\gtv\ $, $v_\fgtv\ $, and $v_\kaduri\ $. 
The resulting vector is the features \mapfgas\ uses for training. 
The training process itself is a straightforward multi-class supervised learning process, 
where every instance is a MAPF problem instance and the label is the fastest algorithm for that instance within our portfolio of algorithms. 
Figure~\ref{fig:mgs-features} illustrates the feature extraction and training process. 
Note that more sophisticated approaches to AS exists in other domains~\cite{kerschke2019automated}, e.g., methods that involve runtime predictions. This is left for future work.  


\section{Experimental Results}
\label{scn:EXPERIMENT}

\begin{figure}
    \centering
    \includegraphics[width=0.23\textwidth]{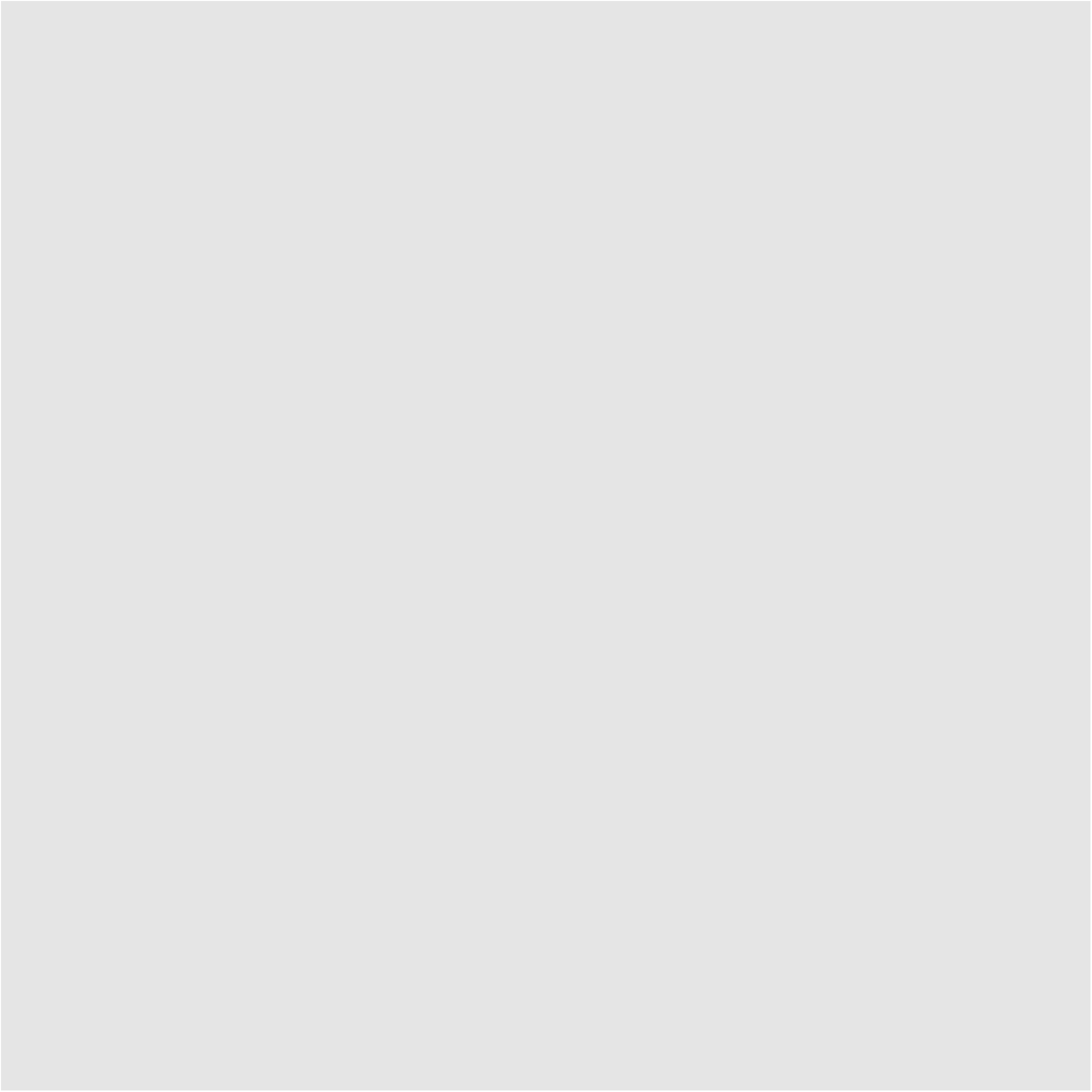}\enspace
    \includegraphics[width=0.23\textwidth]{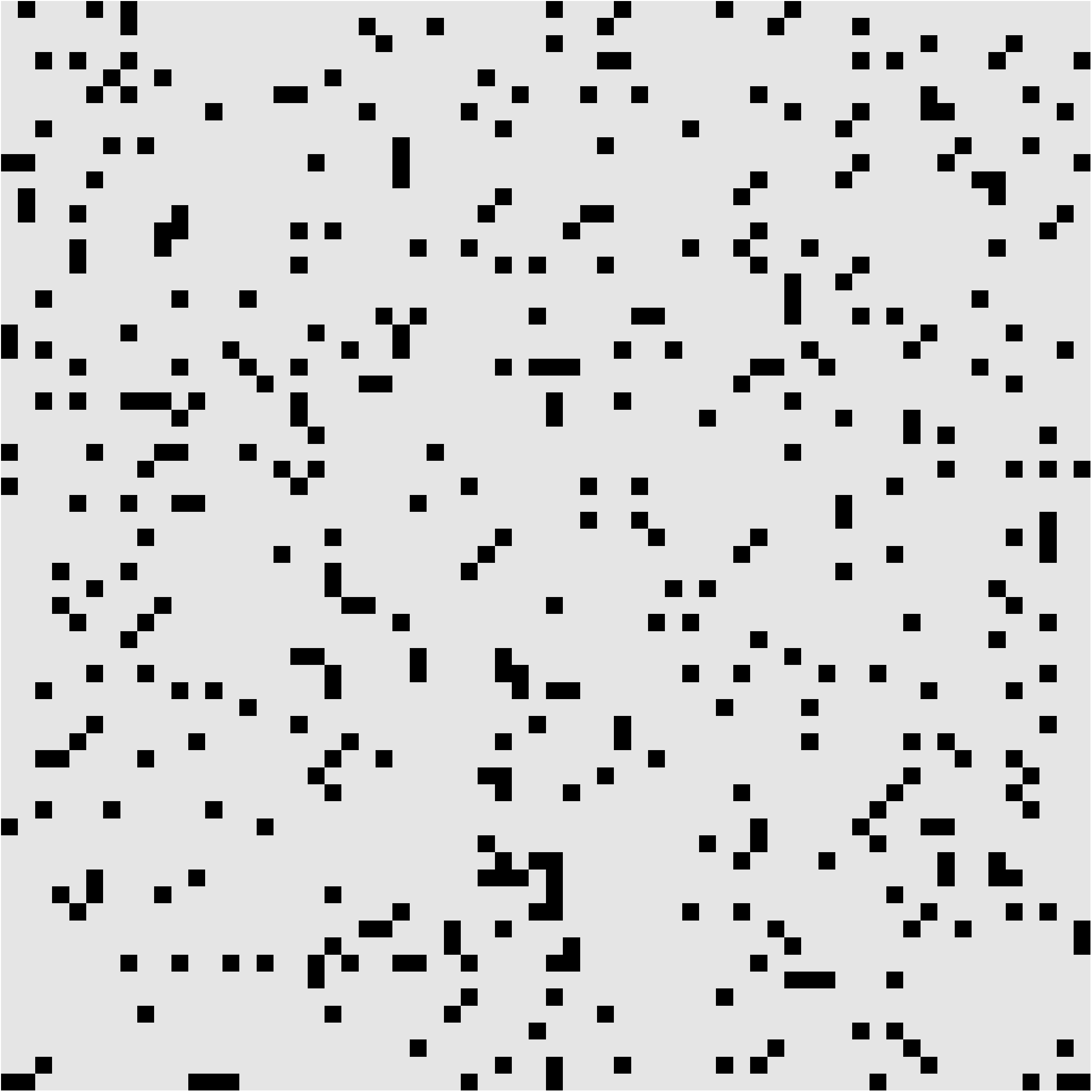}\enspace
    \includegraphics[width=0.23\textwidth]{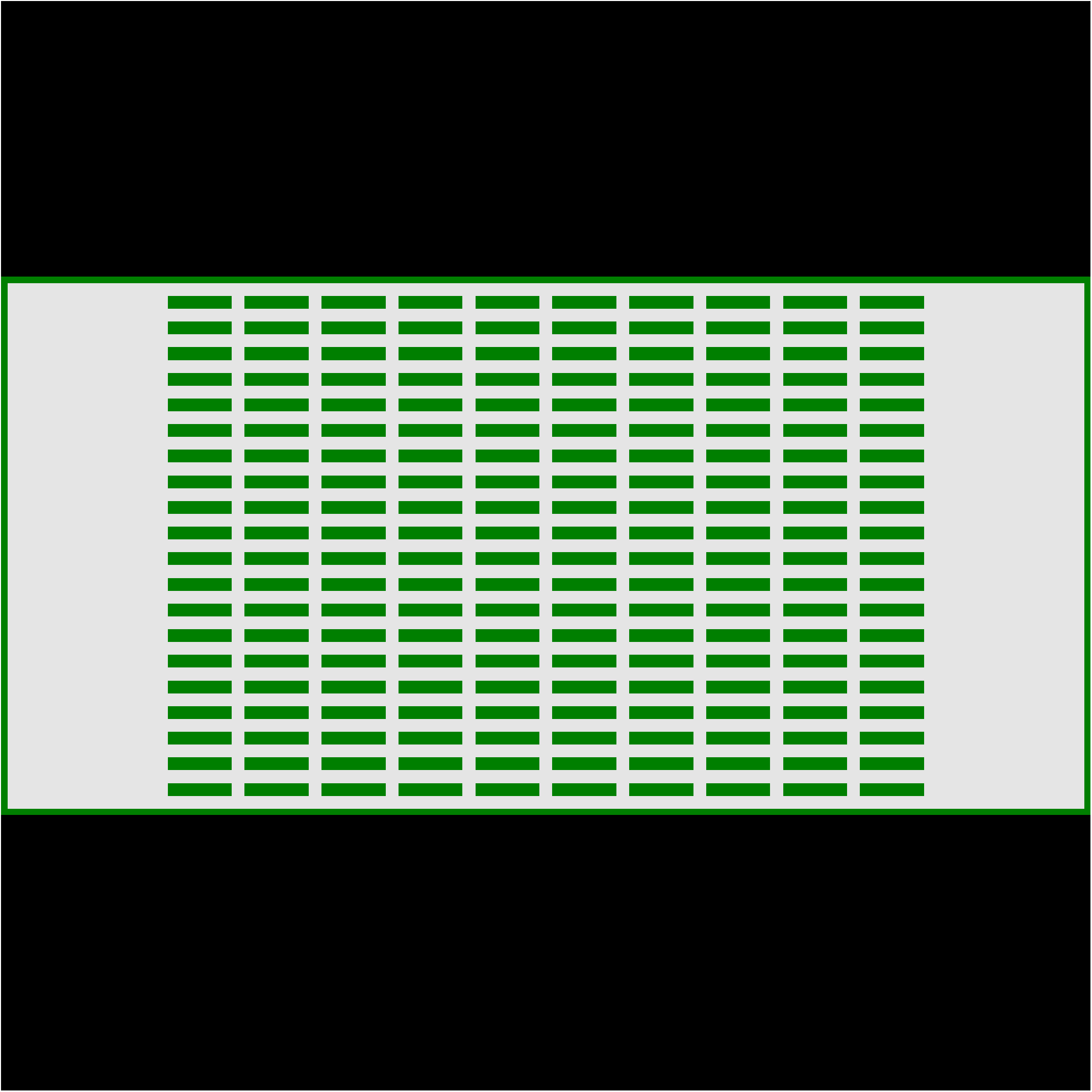}\enspace
    \includegraphics[width=0.23\textwidth]{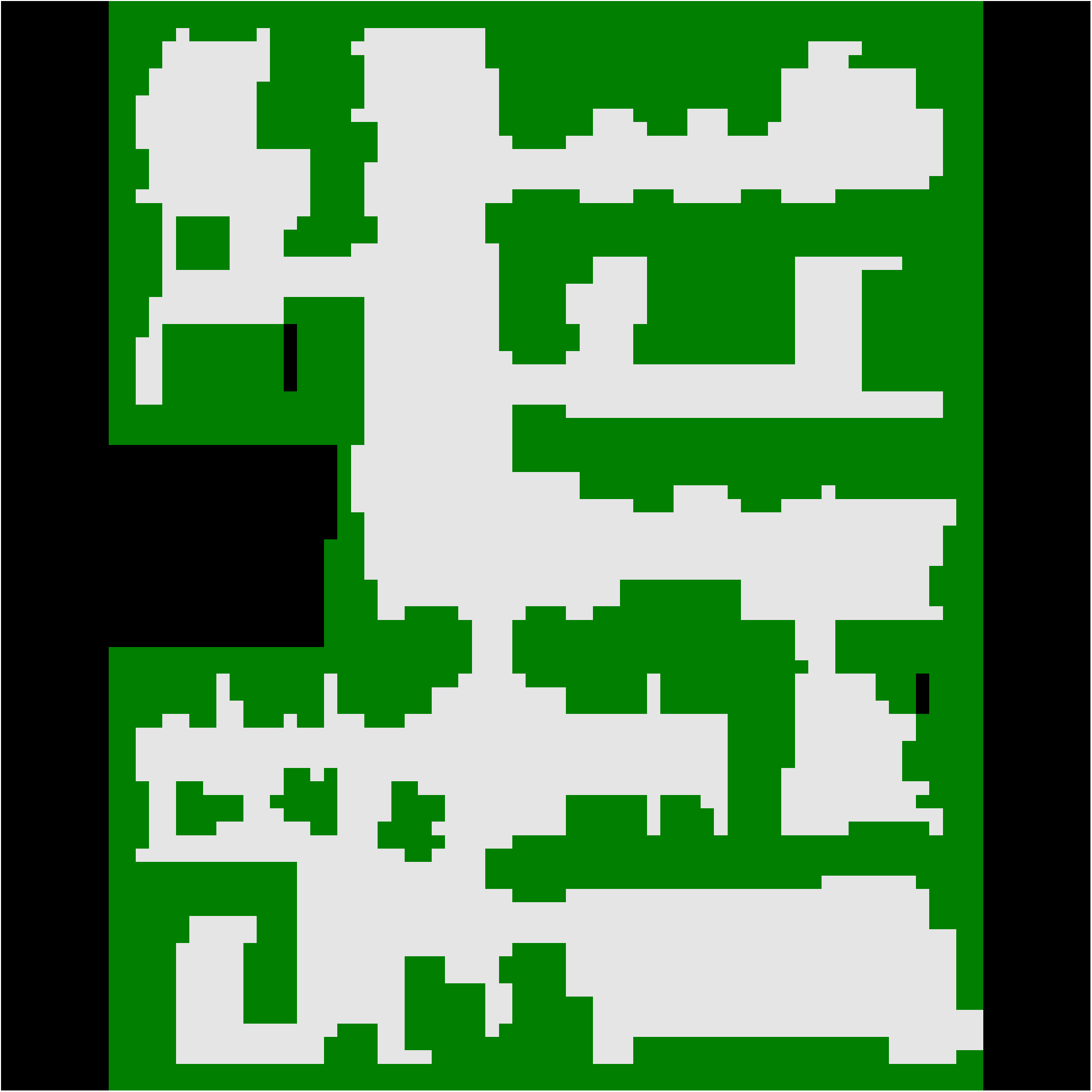}\enspace \\ 
    \includegraphics[width=0.23\textwidth]{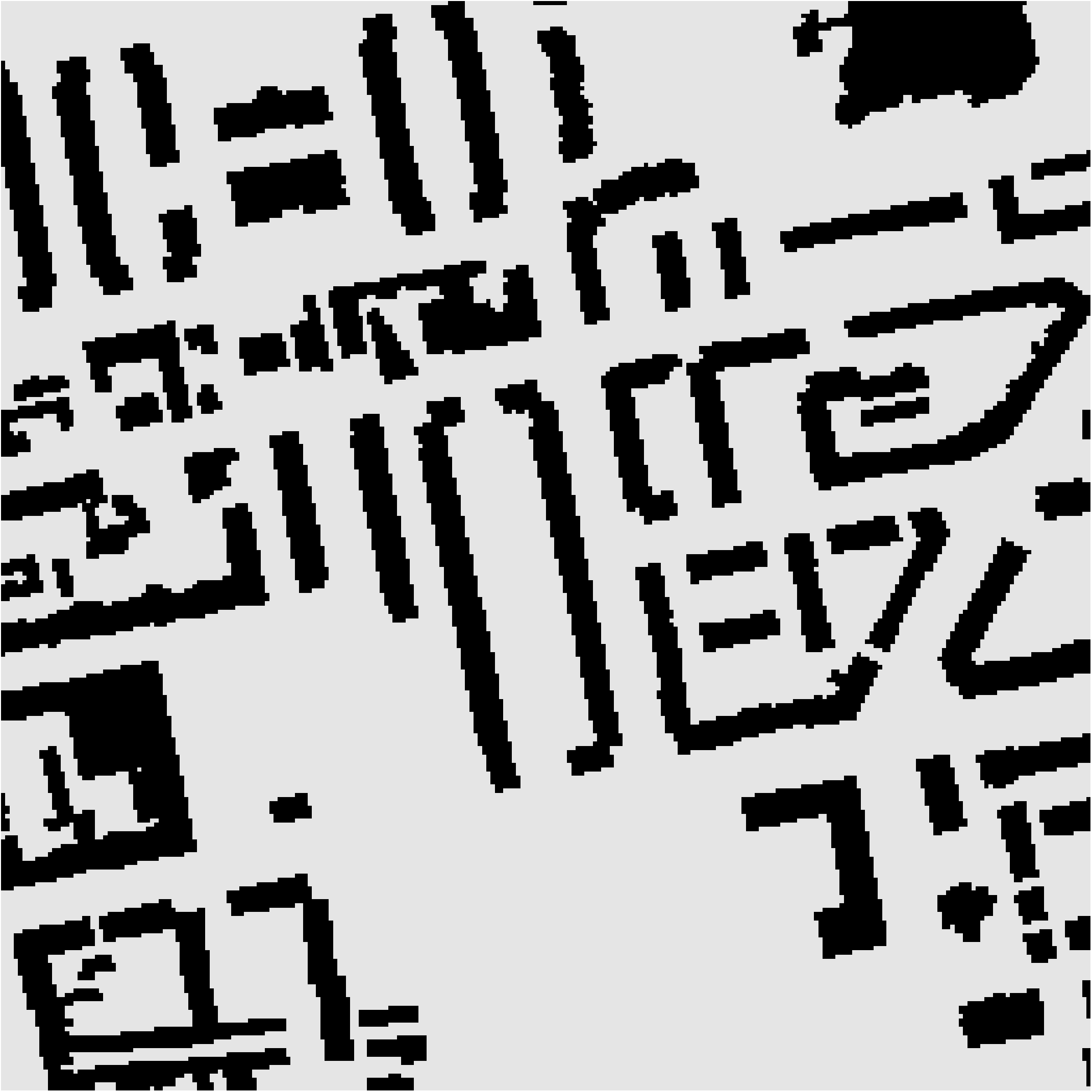}\enspace
    \includegraphics[width=0.23\textwidth]{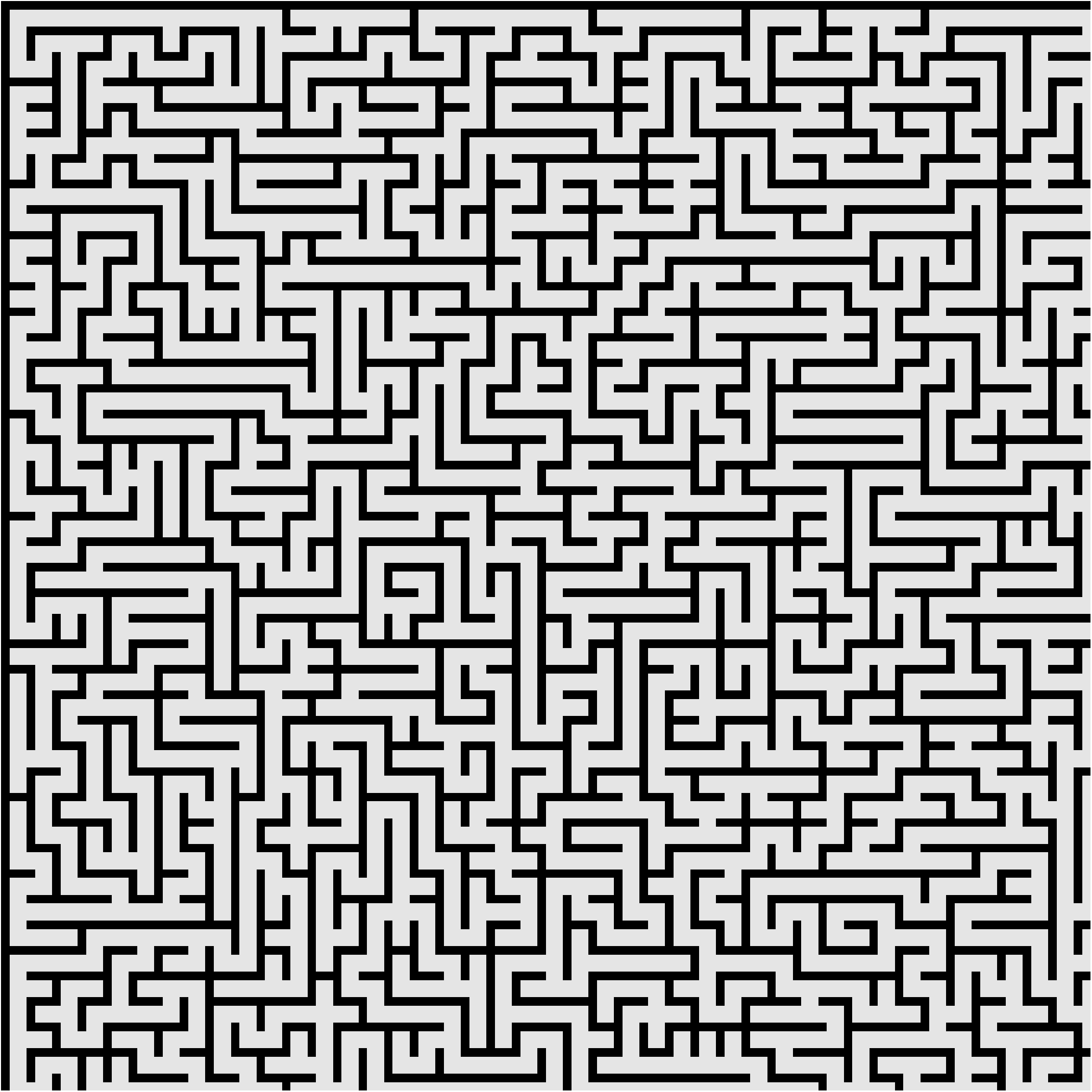}\enspace
    \includegraphics[width=0.23\textwidth]{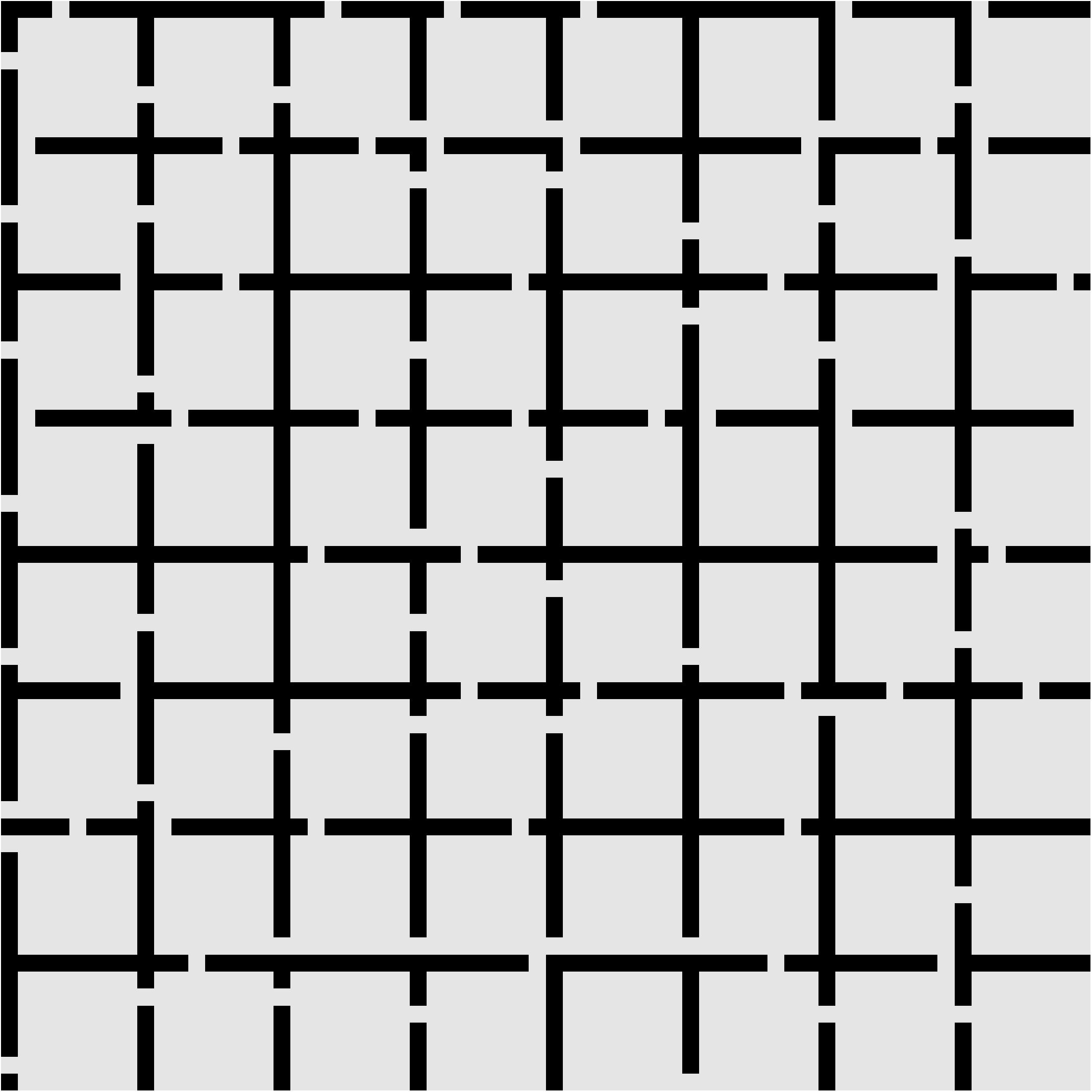}    
    \caption{An example grid from each of the grid types in our benchmark, From left to right: empty, random, warehouse, game, city, maze, and room.}
    \label{fig:grid-types}
\end{figure}

In this section, we present an experimental evaluation of \mapfgas\ on a standard publicly available grid-based MAPF benchmark~\cite{stern2019multi}. 
This benchmark contains 33 grids arranged into seven \emph{grid types}: video
games (denoted as ``game'' grids), city maps (``city''), maze-like
grids (``maze''), grids arranged as rooms with narrow doors
between them (``room''), open grids (``empty''), open grids with
randomly placed obstacles (``random''), and grids that are inspired by the structure of warehouses (``warehouse'').
Figure~\ref{fig:grid-types} shows an example grid from each type.\footnote{The images were taken from the MovingAI repository~\cite{sturtevant2012benchmarks}, which hosts the grid MAPF benchmark we used~\cite{stern2019multi}.}
The benchmark includes \emph{scenario files} for each grid. 
Each scenario file contains source and target locations for as many as 1,000 agents.\footnote{Every scenario file contains 1,000 pairs of start and target locations, except for grids that are too small to occupy that many agents.} 
The scenario files of each grid are grouped into two sets. 
In the first set of scenario files, denoted
\emph{Random}, 
the agents' source and target locations are located
purely randomly. 
In the second set of scenario files, denoted
\emph{Even}, the agents’ source and target locations are evenly distributed in buckets of 10 agents according to the distance between each agent's start and target. 
Only the scenarios from the Even set were used, as they represent a more diverse set of MAPF problems. 

In total, the training sets in our experiments included an average 3,000 instances for each map. The entire training process (across all training set instances) required approximately 10 to 12 hours, including feature creation, training, and hyperparameter tuning. 
The runtime required for feature extraction of a single instance was a single second on average, mostly devoted to running FEATHER. 
The prediction (inference) runtime of our XGBoost model is also extremely fast, since it is not a large deep NN. 
Thus, the overall runtime required to obtain the AS predictions for all our models was negligible compared to the running time of the chosen AS algorithms. 
All experiments were run on a server with an Intel 12th Gen i7-12850HX 2.10 GHz CPU, 64.0 GB RAM, and a 64-bit OS. We are fully committed to making the code, datasets, and results publicly available once the paper is accepted.

\subsection{Experimental Setup}

We performed three sets of experiments, one for each of the AS problem setups defined by Kaduri et al.~\cite{kaduri2021experimental}: in-grid, in-grid-type, and between-grid-type. 
To train and evaluate \mapfgas\ in each setup, we used the publicly available dataset of Kaduri et al.~\shortcite{kaduri2021experimental}, 
which includes results for running a set of optimal MAPF solvers of the entire MAPF benchmark mentioned above. 
Specifically, results for the following optimal MAPF solvers are available in this dataset: ICTS~\cite{sharon2013increasing}, EPEA*~\cite{goldenberg2014enhanced}, SAT-MDD~\cite{surynek2016efficient}, CBSH~\cite{felner2018adding}, and Lazy CBS~\cite{gange2019lazy}. 
A time limit of 5 minutes was imposed for each run. This time limit is common in the Optimal MAPF literature~\cite{sharon2013increasing,sharon2015conflict,goldenberg2014enhanced}. 

\subsubsection{Baselines}
\noindent We compared \mapfgas\ against two baselines: 
\begin{itemize}
    \item \textbf{\kaduri\ .} The AS method by Kaduri, Boyarski and Stern~\cite{kaduri2020algorithm}, which uses only their hand crafted features. 
    \item \textbf{\gtv\ .} The AS method described by Ren et al.\cite{ren2021mapfast}, which uses only the \gtv\ encoding. To extract features from the \gtv\ graph, we used the FEATHER graph embedding.\footnote{This differs from Ren et al., who used Graph2Vec. As explained earlier, Graph2Vec is not a practical method for our problem, since it requires knowing a-priori the graphs to embed.}
\end{itemize}
We do not compare against baselines that always choose the same algorithm, as Kaduri et al.~\shortcite{stern2019multi} already established that such baselines yield poorer results compared to \kaduri\ . Prior work already established that AS with the \kaduri\ features yielded better results 
In addition, we performed an ablation study for \mapfgas\ , and report on results for AS methods that use different subsets of features. Namely, \fgtv\ + \gtv\, 
\kaduri\ + \fgtv\, 
\kaduri\ + \gtv\, 
and \fgtv\ .
Note that \kaduri\ + \gtv\ + \fgtv\ is exactly \mapfgas\ .
For training, we used XGBoost~\cite{chen2016xgboost}, a well-known supervised learning algorithm. 
Preliminary experiments with other learning algorithms, such as Logistic Regression and Random Forest, yielded weaker results. 
The hyperparameters of XGBoost were tuned by performing a  4-fold cross-validation over the training set, for each AS setup.

\subsubsection{Metrics}

\noindent To measure the performance of each AS method, we considered the same metrics as Kaduri et al.~\cite{kaduri2021experimental}. 
\begin{itemize}
    \item \textbf{Accuracy (Acc).} Ratio of instances where the AS method returned the fastest MAPF algorithm. 
    \item \textbf{Coverage (Cov).} Ratio of MAPF instances solved under the 5-minutes time limit. 
    \item \textbf{Runtime (RT).} Average runtime in minutes to solve a single MAPF instance with the selected MAPF solver.\footnote{MAPF instances that were not solved under the 5-minutes time limit were regarded as having a runtime of 5 minutes. This was also done by prior work on AS for MAPF~\cite{kaduri2020algorithm,ren2021mapfast}.}  
\end{itemize}

\Roni{new text}
A limitation of the RT metric is that it gives more weight to MAPF instances that take more time to solve. 
To address this limitation, we propose a novel metric that considers the runtime of an AS method with respect to the runtime of an \emph{Oracle} AS method, which always selects the fastest algorithm for every MAPF instance. 
No practical AS method can perform better than Oracle, which has accuracy and coverage of 1.0, and the smallest possible runtime. We define the \emph{regret} of an AS algorithm on a specific MAPF instance as the percentage of runtime the chosen MAPF instance required over the runtime of the Oracle for that instance.  
Formally, for an AS method $\textit{Alg}$ and a MAPF instance $\Pi$, let $\textit{Alg}(\Pi)$ denote the runtime required to solve $\Pi$ using the MAPF algorithm chosen by $\textit{Alg}$ for $\Pi$. 
Similarly, $\textit{Oracle}(\Pi)$ is the runtime required to solve $\Pi$ using the MAPF algorithm chosen by Oracle. 
The \emph{regret} of $\textit{Alg}$ for $\Pi$ is defined as follows:
\begin{equation}
    \%Rg(Alg,\Pi)=100\cdot \frac{Alg(\Pi)-Oracle(\Pi)}{Oracle(\Pi)}~~~
\end{equation}
\noindent The regret of Oracle is by definition zero, and AS methods with low average regret values are, in general, preferred. 
Note that since all solvers are optimal MAPF solvers, all solvers return solutions of exactly the same cost. Thus, comparing solution quality is redundant. 

\subsection{In-Grid Results}

\begin{table}
\centering
\begin{tabular}{@{}lrrr|rrrr@{}}
\toprule
         & \multicolumn{3}{c}{All}                                                    & \multicolumn{4}{c}{Avg}                                                                                \\ 
Metric   & \multicolumn{1}{c}{Acc} & \multicolumn{1}{c}{Cov} & \multicolumn{1}{c|}{RT} & \multicolumn{1}{c}{Acc} & \multicolumn{1}{c}{Cov} & \multicolumn{1}{c}{RT} & \multicolumn{1}{c}{\%Rg} \\ \midrule
\kaduri{}      & 0.83                    & 0.98                    & 0.549                  & 0.88                    & \textbf{0.99}           & 0.41                   & 12.4                      \\
\gtv{}      & 0.81                    & 0.97                    & 0.589                 & 0.87                    & 0.98                    & 0.45                   & 25.4                      \\
\mapfgas\   & \textbf{0.85}           & \textbf{0.99}           & \textbf{0.514}         & \textbf{0.89}           & \textbf{0.99}           & \textbf{0.40}          & \textbf{9.0} \\ \midrule
Oracle   & 1.00                    & 1.00                    & 0.436                  & 1.00                    & 1.00                    & 0.36                   & 0.0                       \\\midrule
\rowcolor{Gray}
\fgtv{}+\gtv{} & 0.84                    & 0.98                    & 0.531                  & 0.89                    & 0.99                    & 0.41                   & 11.5                      \\
\rowcolor{Gray}
\kaduri{}+\gtv{}  & 0.84                    & 0.98                    & 0.532                  & 0.89                    & 0.99                    & 0.40                   & 9.5                       \\
\rowcolor{Gray}
\kaduri{}+\fgtv{} & 0.85                    & 0.99                    & 0.513                  & 0.89                    & 0.99                    & 0.39                   & 8.1                       \\
\rowcolor{Gray}
\fgtv{}     & 0.84                    & 0.98                    & 0.534                  & 0.88                    & 0.99                    & 0.41                   & 11.8                      \\ \bottomrule
\end{tabular}
\caption{Results for the In-Grid AS setup, averaged over all test problems.} 
\label{tab:in-grid-all}
\end{table}

Table~\ref{tab:in-grid-all} presents the results for the in-grid AS setup experiments. 
The rows correspond to different AS methods, and the columns are the metrics defined earlier. 
The column groups ``All'' and ``Avg'' provide a slightly different way to aggregate the results over all the test problems. 
The results under the ``All'' columns are averages over all test problems, regardless of their grids and grid types. 
This is how most prior work on AS for MAPF aggregated their results. 
The limitation of this aggregation is that some grids in the benchmark are smaller than others, and have fewer MAPF problems defined for them. 
The results under the ``Avg'' columns are averages of averages, where the results of each grid type are averaged separately and only the resulting averages are averaged. This mitigates unwanted to bias stemming from the number of problems in each grid type. 


Consider first the results for \mapfgas\ and our two main baselines, \kaduri{} and \gtv{}.
For each metric, we highlighted the best results among these AS methods in bold. 
As the results clearly show, \mapfgas\ is either on par or better than these baselines on all metrics. 
For example, the accuracy of \mapfgas\ is 0.85 in ``All'' while it is 0.83 and 0.81 for \kaduri\ and \gtv{}, respectively. 
The advantage of \mapfgas\ over \gtv\ is more significant, and more modest compared to \kaduri{}. 
Still advantage is significant, especially in terms of the average regret, which is approximately 25\% smaller than \kaduri\ (9.0 vs. 12.4). 

The results of our ablation study are given in the shaded rows of Table~\ref{tab:in-grid-all}. The results are very similar to \mapfgas, where there is a slight advantage for using the \kaduri\ hand-crafted MAPF-specific features together with graph embedding features (either \kaduri\ + \gtv\ or \kaduri\ + \fgtv\ ). 
This is expected, as a more diverse set of features is expected to be more beneficial. That being said, in some cases some of the algorithm configurations in our ablation study performed as well and even slightly better than the full \mapfgas. Automated methods for algorithm configuration can potentially be used to identify the optimal configuration for a given problem instance. This is beyond the scope of this work.

\subsection{In-Grid-Type and Between-Grid-Type Results}

\begin{table}
\centering
\begin{tabular}{@{}l|rrrr|rrrr@{}}
\toprule
     & \multicolumn{4}{c|}{In-Grid-Type} & \multicolumn{4}{c}{Between-Grid}       \\ 
Metric   & Acc    & Cov    & RT    & \% Rg  & Acc    & Cov    & RT    & \% Rg  \\\midrule
\kaduri{}      & 0.69   & 0.93   & 0.86  & 78.6  & \textbf{0.63}     & 0.87     & 0.99     & 223.1 \\
\gtv{}      & 0.67   & 0.92   & 0.91  & 91.8  & 0.61     & 0.86     & 1.04     & 220.7 \\
 \mapfgas\  & \textbf{0.71}   & \textbf{0.94}   & \textbf{0.80}  & \textbf{67.9}  & \textbf{0.63}     & \textbf{0.89}     & \textbf{0.93}     & \textbf{180.0} \\ \midrule
Oracle   & 1.00   & 1.00   & 0.48  & 0.00   & 1.00     & 1.00     & 0.36     & 0.00   \\ \midrule
\rowcolor{Gray}
\fgtv{}+\gtv{} & 0.69   & 0.93   & 0.84  & 75.8  & 0.62     & 0.88     & 0.97     & 195.8 \\
\rowcolor{Gray}
\kaduri{}+\fgtv{} & 0.70   & 0.94   & 0.81  & 70.5  & 0.64     & 0.89     & 0.93     & 192.5 \\
\rowcolor{Gray}
\kaduri{}+\gtv{}  & 0.70   & 0.94   & 0.82  & 71.6  & 0.65     & 0.88     & 0.94     & 181.8 \\
\rowcolor{Gray}
\fgtv{}     & 0.66   & 0.91   & 0.90  & 87.9  & 0.62     & 0.88     & 0.95     & 176.0 \\ \bottomrule
\end{tabular}
\caption{In-grid-type (left) and Between-grid-type (right) results.}
\label{tab:in-and-between-grid-type}
\end{table}

The first 5 columns (left-to-right) in Table~\ref{tab:in-and-between-grid-type} present the results for the in-grid-type AS setup experiments. 
Similar to Table~\ref{tab:in-grid-all}, the rows are different AS methods and the columns are different metrics, corresponding to the ``Avg'' column family. 
We highlighted in bold the AS method, among \mapfgas\ and our two baselines,  that yielded the best results in each metric. 
As in the in-grid experiments, the advantage of \mapfgas\ over the baselines is clear in all metrics. For example, its average regret is 67.9 while it is 78.60 and 91.80 for \kaduri\ and \gtv, respectively. 
\Roni{New text}
To verify the statistical significance of our results, 
we performed a one-sided $t$ test~\cite{semenick1990tests} to compare \mapfgas{} against the two baseline sets of features, \kaduri{} and \gtv{}. 
This test was done over all grid types, randomizing the selection of test and train MAPF instances for five times. 
The results show that the advantage of \mapfgas{} over \kaduri{} and \gtv{} is significant in terms of accuracy (Acc), with $p$-values of 0.035 and 0.000027, respectively. 
Similarly, the advantage of \mapfgas{} is also significant in terms of coverage, with a $p$ value of 0.007 and 0.00055 for \kaduri{} and \gtv{}, respectively. 
In terms of regret, the advantage of \mapfgas{} over \gtv{} is significant ($p$ value of 0.001) but the advantage of \mapfgas{} over \kaduri{} is not significant, with a $p$ value of 0.397. 
\Roni{End new text}

\begin{table*}[bthp]
\centering
\resizebox{\textwidth}{!}{
\begin{tabular}{@{}ll|rrr|rrr|rrr@{}}
\toprule
                           & AS Setup & \multicolumn{3}{c}{In-Grid}                                                 & \multicolumn{3}{c}{In-Grid-Type}                                            & \multicolumn{3}{c}{Between-Grid}                                            \\ 
Grid-type                  & Metric   & \multicolumn{1}{c}{KBS} & \multicolumn{1}{c}{G2V} & \multicolumn{1}{c}{MAG} & \multicolumn{1}{c}{KBS} & \multicolumn{1}{c}{G2V} & \multicolumn{1}{c}{MAG} & \multicolumn{1}{c}{KBS} & \multicolumn{1}{c}{G2V} & \multicolumn{1}{c}{MAG} \\ \midrule
\multirow{3}{*}{Empty}     & Acc      & 0.88                    & 0.89                    & \textbf{0.90}           & 0.67                    & 0.68                    & \textbf{0.71}           & 0.81                    & 0.78                    & \textbf{0.82}           \\
                           & Cov      & \textbf{1.00}           & \textbf{1.00}           & \textbf{1.00}           & 0.97                    & \textbf{1.00}           & 0.99                    & \textbf{1.00}           & 0.99                    & \textbf{1.00}           \\
                           & \%Rg     & 0.90                     & \textbf{0.10}            & 0.80                     & 3.50                     & \textbf{1.00}            & 1.80                     & 9.50                     & 17.70                    & \textbf{2.50}            \\
\midrule
\multirow{3}{*}{Random}    & Acc      & \textbf{0.91}           & 0.90                    & \textbf{0.91}           & \textbf{0.84}           & 0.79                    & 0.82                    & \textbf{0.78}           & 0.71                    & 0.73                    \\
                           & Cov      & \textbf{1.00}           & \textbf{1.00}           & \textbf{1.00}           & \textbf{1.00}           & 0.99                    & \textbf{1.00}           & \textbf{1.00}           & 0.99                    & \textbf{1.00}           \\
                           & \%Rg     & 3.10                     & 4.10                     & \textbf{2.40}            & 1.40                     & 2.30                     & \textbf{1.10}            & 7.20                    & 54.80                    & \textbf{4.00}            \\
\midrule
\multirow{3}{*}{Warehouse} & Acc      & 0.84                    & 0.84                    & \textbf{0.86}           & 0.64                    & 0.64                    & \textbf{0.67}           & 0.67                    & 0.62                    & \textbf{0.68}           \\
                           & Cov      & 0.99                    & 0.99                    & \textbf{1.00}           & 0.94                    & 0.94                    & \textbf{0.96}           & 0.96                    & 0.88                    & \textbf{0.98}           \\
                           & \%Rg     & 7.9                     & 9.40                     & \textbf{4.70}            & 1.40                     & 1.60                     & \textbf{1.30}            & 32.9                    & 78.10                    & \textbf{27.3}           \\
\midrule
\multirow{3}{*}{Game}      & Acc      & 0.91                    & 0.91                    & \textbf{0.92}           & 0.77                    & 0.72                    & \textbf{0.78}           & \textbf{0.74}           & 0.72                    & 0.65                    \\
                           & Cov      & 0.98                    & 0.98                    & \textbf{0.99}           & \textbf{0.92}           & 0.86                    & \textbf{0.92}           & 0.88                    & \textbf{0.89}           & \textbf{0.89}           \\
                           & \%Rg     & 19.2                    & 21.00                  & \textbf{18.00}           & \textbf{1.40}            & 2.30                     & 1.70                     & 122.0                   & \textbf{113.10}                   & 122.00          \\
\midrule
\multirow{3}{*}{City}      & Acc      & 0.90                    & 0.89                    & \textbf{0.92}           & 0.58                    & 0.57                    & \textbf{0.61}           & 0.53                    & 0.53                    & \textbf{0.58}           \\
                           & Cov      & \textbf{0.99}           & \textbf{0.99}           & \textbf{0.99}           & 0.81                    & \textbf{0.83}           & \textbf{0.83}           & \textbf{0.90}           & 0.86                    & 0.88                    \\
                           & \%Rg     & 10.10                    & 12.40                    & \textbf{7.10}            & 3.10                     & 3.00                     & \textbf{2.90}            & \textbf{136.20}          & 155.60                   & 146.30                   \\
\midrule
\multirow{3}{*}{Maze}      & Acc      & 0.85                    & 0.79                    & \textbf{0.86}           & 0.40                    & 0.44                    & \textbf{0.50}           & \textbf{0.56}           & 0.45                    & 0.47                    \\
                           & Cov      & 0.98                    & 0.93                    & \textbf{0.99}           & \textbf{0.66}           & 0.62                    & \textbf{0.66}           & \textbf{0.71}           & 0.64                    & 0.66                    \\
                           & \%Rg     & 31.90                    & 119.30                  & \textbf{20.60}           & \textbf{8.20}            & 9.60                     & 8.40                     & \textbf{398.8}          & 511.40                   & 483.60                   \\
\midrule
\multirow{3}{*}{Room}      & Acc      & \textbf{0.93}           & 0.90                    & 0.92                    & 0.63                    & 0.50                    & \textbf{0.66}           & 0.32                    & 0.43                    & \textbf{0.49}           \\
                           & Cov      & \textbf{1.00}           & \textbf{1.00}           & \textbf{1.00}           & 0.91                    & 0.83                    & \textbf{0.94}           & 0.65                    & 0.75                    & \textbf{0.81}           \\
                           & \%Rg     & 0.20                     & 2.10                     & \textbf{0.10}            & 2.90                     & 5.00                     & \textbf{2.20}            & 855.60                   & 614.50                   & \textbf{475.00}          \\ \bottomrule 
\end{tabular}
}
\caption{Results for all AS setups, grouped by test grid type.}
\label{tab:by-grid-type}
\end{table*}

The ablation study results show that here too, combining the hand-crafted features of \kaduri\ with either type of graph embedding provides the biggest performance improvement. For example, \kaduri\ with either \gtv\ or \fgtv\ yields 0.70 accuracy and runtime of 0.82 or less while \fgtv\ alone or even with \gtv\ yielded lower accuracy and a higher runtime. It is worth comparing the average regret results here and in the in-grid experiments. While the regret of \mapfgas\ here is 67.90 it is only 9.0 in the in-grid results. This highlights that in-grid-type AS is a significantly harder task the in-grid AS, since it requires generalizing from different grids (although from the same type).

The rightmost 4 columns in Table~\ref{tab:in-and-between-grid-type} present the results for the between-grid-type AS setup experiments. 
The first trend we observe is that the overall results for all algorithms are significantly worse compared to all other AS setups (in-grid and in-grid-type). 
For example, the average regret of \mapfgas\ in the in-grid-type results is 67.9 but it is 180 in between-grid results. Similarly, \mapfgas\ accuracy dropped from 0.71 to 0.63. Recall that the regret and accuracy of \mapfgas\ in the in-grid
was 9.0 and 0.89, respectively. These differences are expected, since in the between-grid experiments, the training set did not include any grid of the tested type, which makes the classification problem significantly harder.

In terms of the comparison with our baselines, the general trend we observed so far continues in the between-grid setup: \mapfgas\ is either on par or better than the baselines in all metrics. 
For example, its average runtime and regret are 0.93 and 180 while it is 0.99 and 220.7 for the next best baseline, respectively. 
The ablation study results are less conclusive in this setup. 
In terms of accuracy, we still see the benefit in combining \kaduri\ features with graph embedding features over only using graph embedding features. However, the lowest average regret is achieved when only using \mapfgas\ . In fact, some subsets of \mapfgas\ features actually outperform \mapfgas\ on some metrics. For example, using only \fgtv\ yields lower accuracy than the full \mapfgas\ but a slightly lower average regret (176 vs. 180). However, these differences are relatively small.

\subsection{Grid Types Analysis}
Table~\ref{tab:by-grid-type} provides a deeper insight into the results of our in-grid, in-grid-type, and between-grid experiments. 
Here, the results --- accuracy, coverage, and regret --- are grouped by grid types. 
For example, the results in the row ``City'' show the average results over test problems that are on grids of type ``City''. 
The rows correspond to the test grid type, and the columns correspond to the AS setup and evaluated algorithm. 
While \mapfgas\ is still, in general, the best-performing algorithm, in some grid types and metrics it is not, especially in the between grid setup. 
This is most evident in the results for Maze grids, where the accuracy, coverage, and regret of \kaduri\ --- 0.56, 0.71, and 398.8, respectively --- are better than the corresponding results of \mapfgas\, which were 0.47, 0.66, and 483.6. 
A possible explanation for this is that MAPF solutions in Maze grids are significantly different from MAPF solutions in grids from other grid types. Maze grids have many narrow corridors, and consequently avoiding conflicts between agents often requires one agent to make a large detour. 

Interestingly, \mapfgas\ performs well on City grids in the in-grid AS setup. This suggests it is able to learn how to act in such grids, if they are given to it for training. 
The in-grid results for Maze grids provide another interesting insight when comparing the results of \mapfgas\ and \gtv\ . As can be seen, the difference in this grid type between \mapfgas\ and \gtv\ is most pronounced: the accuracy, coverage, and regret of \gtv\ are 0.79, 0.93, and 119.3, respectively, while they are 0.86, 0.99, and 20.6 for \mapfgas\ .  
The poor performance of \gtv\ in Maze grids is understandable: in such grids agents often follow paths that are significantly different from their shortest paths to avoid collisions. Such paths are not encoded by \gtv\ . 

\subsection{Feature Importance}


\begin{figure*}[tbhp]
    \centering
    \includegraphics[width=0.59\columnwidth]{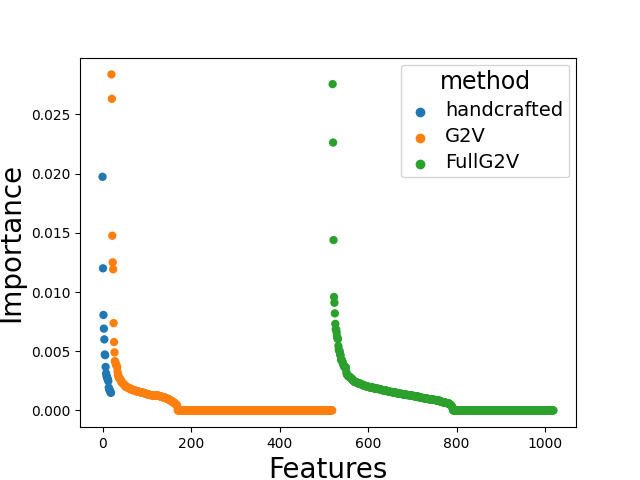}
    \includegraphics[width=0.59\columnwidth]{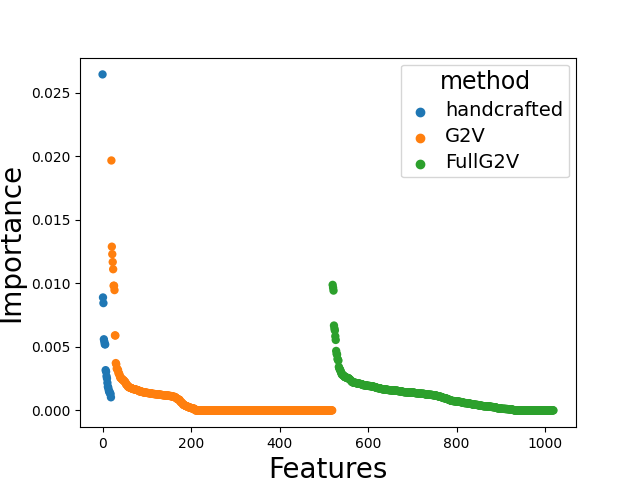}
    \includegraphics[width=0.59\columnwidth]{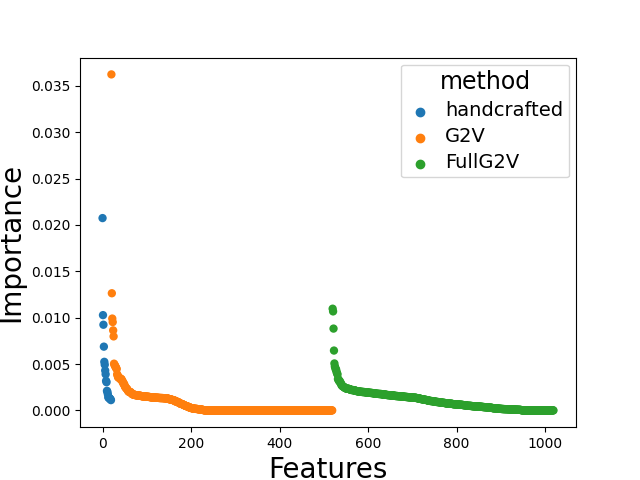}
    \caption{Feature importance for the XGBoost model created for \mapfgas\, in the in-grid (top), in-grid-type (middle), and between-grid (bottom) setups. } 
    \label{fig:model_coefs}
\end{figure*}

Figure \ref{fig:model_coefs} plots a feature importance analysis performed on the prediction models created by \mapfgas\ for each of the AS setups. 
The importance of the different features was measured by the coefficient learned for it by the learning algorithm. 
The features are listed on the x-axis, where the first 20 features are the \kaduri\ features, the next 500 features are the \gtv\ features, and the last 500 features are the \fgtv\ features. Within each feature family (\kaduri\ ,\gtv\ , and \fgtv\ ), the features are sorted by their importance. 

Our main observation is that each feature family has some features that are important. This suggests that there is useful information in all types of features, which corresponds to the successful results of \mapfgas\ . 
Another observation is that each feature family includes features whose importance is close to zero. This suggests that applying feature selection method based on weights may be effective. 
When comparing the different AS setups, it seems that as the AS setup becomes more challenging, the set of features that are important is reduced, and the importance of more features is close to zero. This may suggest that additional types of features may be needed to get better results for the harder AS setups. 

\section{Conclusion and Future Work}
We proposed \mapfgas\ , the first practical approach to optimal MAPF algorithm selection based on graph embedding. 
\mapfgas\ uses two encodings of the MAPF problem as a graph, one that encodes the entire graph and one that encodes only the shortest paths and their immediate vicinity. 
To work efficiently on new MAPF problems, \mapfgas\ utilizes a modern graph embedding algorithm that does not need a-priori training. 
\mapfgas\ also uses hand-crafted MAPF-specific features, as suggested by prior work. 
The combination of graph-embedding features and hand-crafted features leads to strong state-of-the-art AS for optimal MAPF. 
In an extensive set of experiments on a standard benchmark, we showed that \mapfgas\  outperforms existing baselines almost always. 
In the easiest AS setup, In-Grid, the advantage of \mapfgas\ is limited. 
However, in the more challenging AS setups, In-Grid-Type and Between-Grid, the advantage of \mapfgas\ is more significant, with more than 20\% advantage in terms of average regret. 
Our results also highlight that the between-grid AS setup is particularly challenging, and can be the focus of future work. 
Another important future work is to combine our graph-embedding based AS model with image-based AS models, such as MAPFASTER~\cite{alkazzi2022mapfaster}.



\bibliographystyle{elsarticle-num} 
\bibliography{library}





\end{document}